\documentclass[conference]{IEEEtran}
\IEEEoverridecommandlockouts

\usepackage{cite}
\usepackage{amsmath,amssymb,amsfonts}
\usepackage{algorithmic}
\usepackage{graphicx}
\usepackage{textcomp}
\usepackage{xcolor}
\usepackage{booktabs}
\usepackage{url}
\usepackage{float}
\usepackage{caption}
\usepackage[hidelinks]{hyperref}

\usepackage[hidelinks]{hyperref}

\def\BibTeX{{\rm B\kern-.05em{\sc i\kern-.025em b}\kern-.08em
    T\kern-.1667em\lower.7ex\hbox{E}\kern-.125emX}}

\begin{document}

\title{\fontsize{16}{18}\selectfont
Automated Plant Disease and Pest Detection System Using Hybrid Lightweight CNN–MobileViT Models for Diagnosis of Indigenous Crops
}

\author{
\IEEEauthorblockN{Tekleab G. Gebremedhin\textsuperscript{*}, Hailom S. Asegede, Bruh W. Tesheme, Tadesse B. Gebremichael, Kalayu G. Redae}
\IEEEauthorblockA{\textit{Department of Computer Science \& Engineering and Department of Information Technology}\\
\textit{Mekelle University - Mekelle Institute of Technology, Ethiopia}\\
Email: tekleab.gebremedhin@singularitynet.io}
}

\maketitle
\begingroup
\renewcommand\thefootnote{\*}
\footnotetext{Corresponding author. The source code and dataset are publicly available at: \url{https://github.com/Tekleab15/Automated_plant_disease_and_pest_detection_system}}
\renewcommand\thefootnote{\dag}
\footnotetext{A shorter version of this work was presented at the International Conference on Postwar Technology for Recovery and Sustainable Development in Mekelle, Tigray, in Feb.\  2025.}
\endgroup

\begin{abstract}
\normalfont Agriculture supports over 80\% of the population in the Tigray region of Ethiopia,
where infrastructural disruptions limit access to expert crop disease diagnosis.
We present an offline-first detection system centered on a newly curated indigenous
cactus-fig (\textit{Opuntia ficus-indica}) dataset consisting of 3,587 field images
across three core symptom classes. Given deployment constraints in post-conflict,
edge environments, we benchmarked three mobile-efficient architectures:
a custom lightweight CNN, EfficientNet-Lite1, and the CNN-Transformer hybrid
MobileViT-XS.

While the broader system contains independent modules for potato, apple, and corn,
this study isolates cactus-fig model performance to evaluate attention sensitivity
and inductive bias transfer on indigenous morphology alone.

Results establish a Pareto trade-off: EfficientNet-Lite1 achieved 90.7\% test accuracy,
our lightweight CNN reached 89.5\% with the most favorable deployment profile
(42\,ms inference latency, 4.8\,MB model size), and MobileViT-XS delivered
97.3\% mean cross-validation accuracy, showing that MHSA-based global reasoning
disambiguates pest clusters from 2D fungal lesions more reliably than local-texture CNN
kernels. The ARM-compatible models are deployed inside a Tigrigna and Amharic-localized
Flutter application supporting fully offline inference on Cortex-A53-class devices,
strengthening inclusivity for food-security-critical diagnostics.
\end{abstract}


\section{Introduction}
\label{sec:intro}

Agriculture is critical to livelihoods in Ethiopia’s Tigray region, a post-conflict environment
where damaged infrastructure disrupts expert crop disease diagnosis and timely advisory
interventions \cite{fao2022}. The drought-resilient cactus-fig (\textit{Opuntia ficus-indica}),
locally known as “Beles,” plays a dual socio-economic role: as a seasonal buffer crop
supporting food availability during pre-harvest scarcity and as an industrial raw material
used in local manufacturing sectors \cite{opuntia_pest2023}. 

Ecologically unique to the highland agro-system, cactus-fig acts as a strategic buffer crop
against seasonal food scarcity, providing nutrition when alternative yields are depleted.
A major production threat is the invasive cochineal insect (\textit{Dactylopius coccus})
and fungal rots, which form visually ambiguous symptoms (white wax clusters vs. mildew-like
textures) capable of causing large-scale yield devastation if not detected early
\cite{opuntia_pest2023, maize_cnn2021}.

Historically, disease diagnosis relied on manual inspection by agricultural extension workers. However, the recent conflict in the region has severely damaged infrastructure and disrupted the human advisory supply chain, leaving millions of farmers isolated from expert support. In this context, automated, offline-first diagnostic tools are not merely a convenience but a humanitarian necessity.

Recent advances in Computer Vision have enabled automated disease recognition, with Convolutional Neural Networks (CNNs) achieving high accuracy on global datasets like PlantVillage \cite{mohanty2016}. 

However, deploying these models for indigenous highland crops presents two specific scientific challenges:
\begin{enumerate}
    \item \textbf{The Data Gap:} Xerophytic crops like Cactus-fig are morphologically distinct from the broad-leaf crops (e.g., Apple, Corn) found in standard datasets. Their cladodes (pads) exhibit waxy surfaces, glochids (spines), and irregular 3D structures that confound standard pre-trained models.
    \item \textbf{The Architectural Dilemma:} Distinguishing between fungal lesions (2D surface textures) and cochineal infestations (3D clustered pests) requires a model that captures both \textit{local} texture details and \textit{global} context. Standard CNNs excel at local features but lack the global receptive field to identify scattered pest clusters. Conversely, Vision Transformers (ViTs) capture global context but are often too computationally heavy for the low-end hardware found in rural Ethiopia.
\end{enumerate}

To bridge this gap, this paper presents an \textbf{Automated Plant Disease and Pest Detection System} built upon a novel, field-verified dataset of indigenous crops. We propose a hybrid approach, benchmarking a custom \textbf{Lightweight CNN} (optimized for extreme efficiency) against the \textbf{MobileViT-XS} architecture, a CNN-ViT hybrid that combines the inductive bias of convolutions with the self-attention of Transformers.

Our specific contributions are:
\begin{itemize}
    \item \textbf{Indigenous Dataset Construction:} We introduce a dataset of 3,587 annotated images of \textit{Opuntia ficus-indica}, capturing real-world field variability (dust, shadows, mixed infections) often absent in laboratory datasets.
    \item \textbf{Hybrid Architecture Benchmarking:} We rigorously evaluate the trade-off between Inductive Bias (CNN) and Self-Attention (ViT). We demonstrate that the hybrid MobileViT-XS architecture achieves SOTA accuracy (97.3\%) by effectively resolving visual ambiguities that mislead pure CNNs.
    \item \textbf{Offline-First Deployment:} We deploy the optimized models in a Tigrigna and Amharic-localized Flutter application capable of running on ARM Cortex-A53 devices without internet connectivity, ensuring equitable access for resource-constrained farming communities.
\end{itemize}

The remainder of this paper is organized as follows: Section~\ref{sec:background} provides biological and technical background. Section~\ref{sec:methodology} details the dataset and hybrid model architectures. Section~\ref{sec:experiments} presents the comparative results and explainability analysis. Section~\ref{sec:conclusion} concludes with implications for food security.

\section{Background}
\label{sec:background}

\subsection{Plant Pathology and Visual Symptom Variability}
Plant diseases manifest through complex visual cues such as chlorosis, necrotic spots, pustules, and wilting. From a computer vision perspective, these symptoms present a significant challenge due to their high intra-class variance and inter-class similarity. Symptoms are often non-uniform, multi-stage, and visually confounded by environmental factors such as dust, nutrient deficiencies, or physical damage.

The target crop, \textit{Opuntia ficus-indica} (Cactus-fig), introduces unique morphological constraints. Unlike the planar leaf structures of broad-leaf crops (e.g., maize, apple), cactus cladodes (pads) are voluminous, waxy, and covered in glochids (spines). The primary pest threat, the cochineal insect (\textit{Dactylopius coccus}), creates white, cottony wax clusters that can be visually indistinguishable from certain fungal mildews to standard texture-based classifiers. Discriminating between these 3D pest clusters and 2D fungal lesions requires a model capable of understanding both local texture and global geometric context.

\subsection{Convolutional Neural Networks (CNNs)}
CNNs have established themselves as the standard for agricultural image analysis due to their ability to learn hierarchical feature representations. A typical CNN is composed of stacked convolutional layers that extract local features (edges, textures) and pooling layers that introduce spatial invariance.

Formally, a convolutional layer computes a feature map $\mathbf{F}$ from an input $\mathbf{X}$ using a learnable kernel $\mathbf{W}$ and bias $b$:
\begin{equation}
\mathbf{F}_{i,j} = \sigma \left( \sum_{m} \sum_{n} \mathbf{W}_{m,n} \cdot \mathbf{X}_{i+m, j+n} + b \right)
\end{equation}
where $\sigma$ is a non-linear activation function (e.g., ReLU). 
\begin{figure}
    \centering
    \includegraphics[width=1\linewidth]{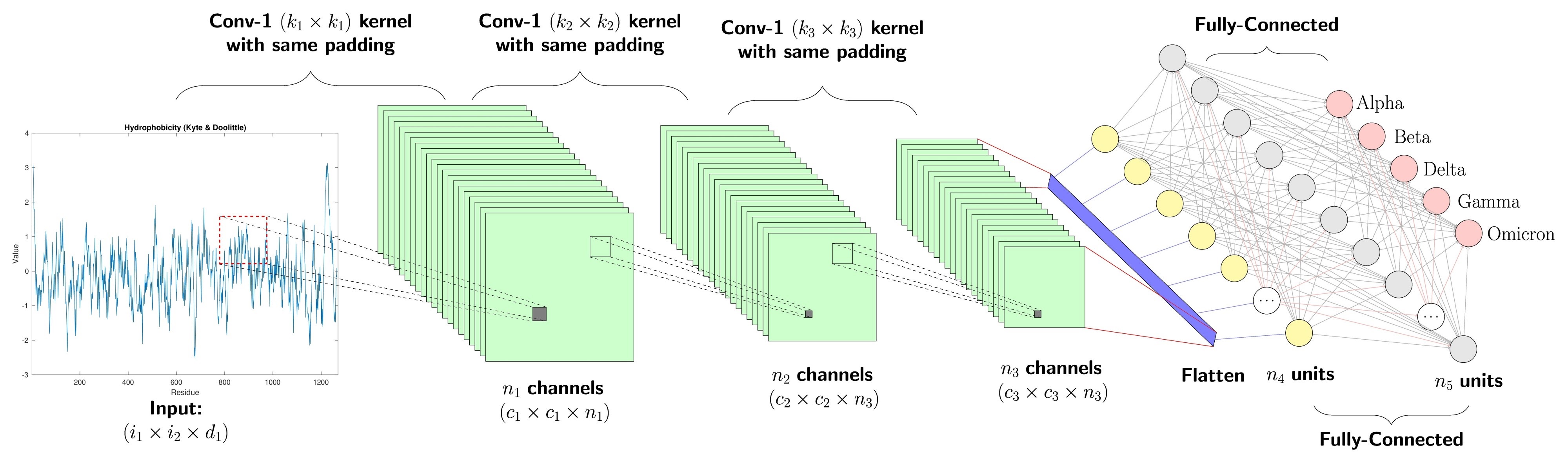}
    \caption{CNN Architecture}
\end{figure}
While CNNs excel at extracting local patterns, their limited receptive field can struggle to capture long-range dependencies, such as the spatial distribution of a pest infestation across a large cladode.

\subsection{Vision Transformers and Self-Attention}
To address the limitations of local receptive fields, Vision Transformers (ViTs) adapt the self-attention mechanism from Natural Language Processing to image data. Unlike CNNs, which process pixels in fixed neighborhoods, Transformers treat an image as a sequence of patches and compute the relationship between every patch pair simultaneously.

\begin{figure}
    \centering
    \includegraphics[width=1\linewidth]{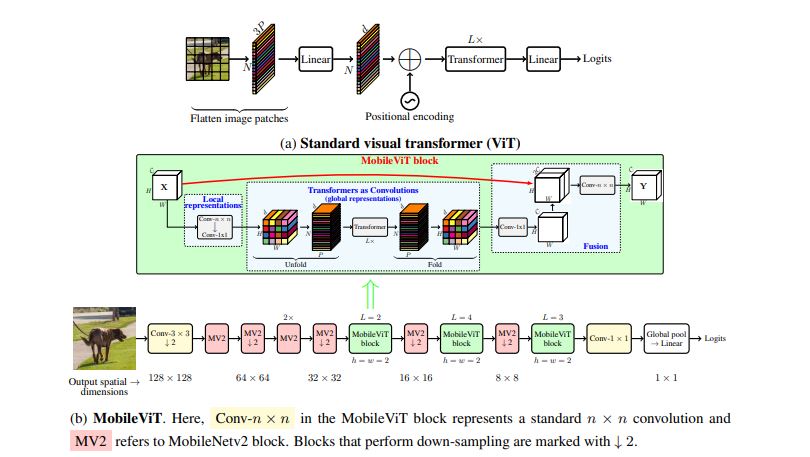}
    \caption{MobileVIT Architecture \cite{mobilevit2022}} 
\end{figure}

The core mechanism is Multi-Head Self-Attention (MSA), which allows the model to weigh the importance of different image regions globally. For a given input sequence $\mathbf{Z}$, the attention output is computed as:
\begin{equation}
\text{Attention}(\mathbf{Q}, \mathbf{K}, \mathbf{V}) = \text{softmax}\left(\frac{\mathbf{Q}\mathbf{K}^T}{\sqrt{d_k}}\right)\mathbf{V}
\end{equation}
where $\mathbf{Q}, \mathbf{K}, \mathbf{V}$ represent the Query, Key, and Value matrices projected from the input. This global context awareness makes ViTs particularly effective at distinguishing complex, scattered pathologies like cochineal clusters from background noise. MobileViT, the architecture used in this study, creates a hybrid by embedding this global processing block inside a lightweight CNN backbone to retain efficiency.

\subsection{Domain Shift in Field Settings}
A critical barrier to deploying AI in agriculture is \emph{domain shift}, where models trained on controlled laboratory images fail in real-world field conditions. Field images introduce uncontrolled variables such as harsh lighting shadows, occlusion by other plants, and background clutter (soil, sky, weeds). For xerophytic crops, the high reflectivity of the waxy cuticle under direct sunlight creates specular highlights that can be misclassified as lesions. This study explicitly addresses domain shift by training on a dataset captured entirely in uncontrolled field environments.

\subsection{Edge AI and Model Quantization}
Deploying deep learning models in rural Tigray requires overcoming severe hardware constraints. Farmers typically rely on low-end Android devices with limited RAM, battery life, and computational power (e.g., ARM Cortex-A53 processors). Cloud-based inference is often unfeasible due to intermittent internet connectivity.

To enable offline execution, we employ \emph{Post-Training Quantization} (PTQ). This process maps high-precision 32-bit floating-point weights ($W_{fp32}$) to lower-precision 16-bit floats ($W_{fp16}$) or 8-bit integers:
\begin{equation}
W_{q} = \text{round}\left( \frac{W_{fp32}}{S} \right) + Z
\end{equation}
where $S$ is a scaling factor and $Z$ is the zero-point. This compression reduces model size and inference latency with minimal degradation in accuracy, making complex architectures like MobileViT deployable on edge devices.

\subsection{Explainability (XAI) in Agriculture}
Trust is a prerequisite for technology adoption in agrarian communities. Black-box AI models offer little insight into their decision-making process. To bridge this gap, we integrate Explainable AI (XAI) techniques such as Local Interpretable Model-agnostic Explanations (LIME). LIME generates saliency maps that highlight the specific image regions driving a prediction. By visualizing that the model is focusing on the pest or lesion rather than background artifacts, we provide agronomic validation that builds user confidence in the automated diagnosis.

\section{Related Work}
\label{sec:related}

The automation of plant disease diagnosis has evolved from hand-crafted feature extraction to end-to-end deep learning. This section reviews the trajectory of these technologies and identifies critical gaps regarding indigenous crops and resource-constrained deployment.

\subsection{Deep Learning in Global Plant Pathology}
The foundational work by Mohanty et al. \cite{mohanty2016} established the viability of Convolutional Neural Networks (CNNs) for crop disease diagnosis, achieving nearly 99\% accuracy on the PlantVillage dataset. Subsequent studies optimized architectures such as ResNet and VGG to improve robustness against lighting variations \cite{too2019}.
However, a systematic review of agricultural AI in Ethiopia \cite{ethiopia_review2025} reveals a persistent bias: models are predominantly trained on globally commercial crops (maize, tomato, apple) with broad planar leaves. These models generalize poorly to xerophytic crops like \textit{Opuntia ficus-indica}, whose morphology (voluminous cladodes, glochids) and pathology (3D cochineal clusters) differ fundamentally from leaf spots \cite{opuntia_pest2023}.

\subsection{Hybrid Architectures and Transformers}
To address the receptive field limitations of CNNs, researchers have increasingly adopted Vision Transformers (ViTs). Recent hybrid architectures like \textbf{MobileViT} \cite{mobilevit2022} and \textbf{PMVT} \cite{pmvt2023} merge the inductive bias of convolutions with the global awareness of self-attention.
In the specific domain of cactus pathology, Berka et al. \cite{cactivit2023} introduced \textbf{CactiViT}, a Transformer-based network designed for diagnosing cactus cochineal infestation. While their work demonstrated the efficacy of attention mechanisms for this crop, it primarily targeted high-end smartphone deployment. This leaves a gap for architectures optimized for the legacy hardware prevalent in post-conflict zones, where computational efficiency is as critical as accuracy.

\subsection{Edge AI and Offline Deployment}
For deployment in the Global South, efficiency is paramount. Architectures like \textbf{EfficientNet} \cite{efficientnet2019} and \textbf{L-Net} \cite{lnet2025} utilize depth-wise separable convolutions to reduce parameter counts. Chowdhury et al. \cite{mobile_system2021} demonstrated that model quantization could enable offline inference on mobile devices.
Despite these advances, a trade-off remains: ultra-lightweight CNNs often sacrifice the semantic understanding required to distinguish complex pest clusters from physical scarring. Furthermore, few systems incorporate Explainable AI (XAI) techniques like LIME \cite{lime2016} directly into the mobile workflow to build user trust in rural communities.

\subsection{Gap Analysis}
Most plant pathology models prioritize global commercial crops using convolution-dominated
benchmarks and static leaf morphology \cite{mohanty2016, too2019}.  
Transformer-based vision models show strong separation but incur \emph{O($N^2$)} self-attention
latency and RAM pressure, limiting fully offline use on low-end devices \cite{cactivit2023, pmvt2023}.  
Lightweight CNNs offer low latency yet fail to encode geometric correlation needed to
separate 3D cochineal clusters from 2D scars or fungal rot on cactus cladodes
\cite{efficientnet2019, atila2021}.  
No prior work benchmarks \emph{indigenous cactus-fig field imagery} with failure-mode
isolation and transfer-leakage prevention.  
This study evaluates a 1.2M-parameter CNN vs. a fine-tuned MobileViT-XS hybrid,
measuring accuracy, latency, and size without cross-crop knowledge leakage.

\section{Methodology}
\label{sec:methodology}

This section details the dataset curation, preprocessing pipeline, and the architectural design of the two competing models: a custom lightweight CNN (efficiency-focused) and the MobileViT-XS hybrid (accuracy-focused). All design choices are motivated by the dual constraints of high diagnostic precision and deployment on low-end mobile hardware in offline environments.

\subsection{Dataset Curation}
\label{subsec:dataset}

\subsubsection{Multi-Crop Backbone}
To ensure the models learn robust, generalized vegetative features, we aggregated a background dataset of 26,394 images from public repositories (e.g., PlantVillage), spanning tomato, potato, maize, and apple crops. This pre-training step stabilizes the early layers of the feature extractors before fine-tuning on the target domain.

\subsubsection{Indigenous Cactus-Fig Dataset}
The core contribution of this work is the curation of a domain-specific dataset for
\textit{Opuntia ficus-indica} (Beles), collected at Mekelle University and Adigrat University
experimental sites. The full dataset has been open-sourced and is accessible via the project
repository and Kaggle. The dataset consists of 3{,}587 high-resolution field images categorized
into three classes:
\begin{itemize}
    \item \textbf{Affected (1,500 images):} Cladodes exhibiting symptoms of cochineal infestation
    (white waxy clusters) or fungal rot (necrotic lesions).
    \item \textbf{Healthy (1,500 images):} Asymptomatic cladodes with natural variations in color and texture.
    \item \textbf{No Cactus (587 images):} Background images (soil, sky, weeds) to train the model
    to reject non-plant inputs.
\end{itemize}

\begin{center}
    \includegraphics[width=0.9\linewidth]{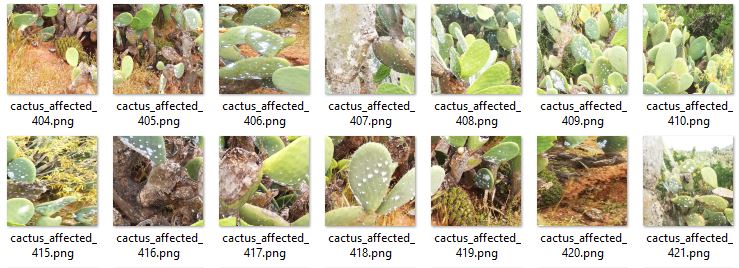}
    \captionof{figure}{Example of affected cactus-fig cladode showing cochineal clusters and fungal lesions.}
    \label{fig:cactus_affected_sample}
\end{center}

\begin{center}
    \includegraphics[width=0.9\linewidth]{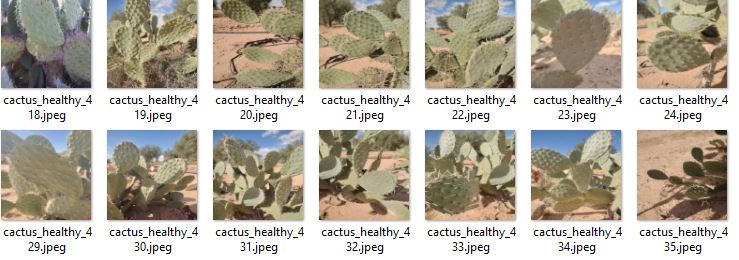}
    \captionof{figure}{Example of healthy cactus-fig cladode with no disease or pest symptoms.}
    \label{fig:cactus_healthy_sample}
\end{center}

\subsection{Preprocessing and Augmentation}
Input images are resized to $256 \times 256$ pixels and normalized to the unit interval $[0,1]$. To mitigate overfitting given the limited size of the indigenous dataset, we employ a \textbf{Sanitized Validation Protocol}: geometric augmentations are applied \textit{only} to the training split, while validation is performed on clean, static images.
The augmentation function $\mathcal{A}(X;\theta)$ is defined as:
\begin{equation}
    \tilde{X} = \mathcal{A}(X;\theta) = T_{\text{rot}}^{\theta_r} \cdot T_{\text{zoom}}^{\theta_z} \cdot T_{\text{flip}}^{\theta_f}(X)
\end{equation}
where $\theta$ represents random parameters for rotation ($\pm 20^\circ$), zoom ($\pm 15\%$), and horizontal flipping.

\subsection{Model Architectures}
We investigate two distinct architectural paradigms to identify the optimal trade-off for edge deployment.

\subsubsection{ Custom Lightweight CNN (Efficiency-First)}
Designed for ultra-low latency on legacy devices, this model consists of three sequential convolutional blocks. Each block $k$ performs feature extraction defined by:
\begin{equation}
    F_{k}(i,j) = \text{Pool}\left(\sigma\left(\sum_{m,n} W_{k}(m,n) \cdot X(i+m,j+n) + b_{k}\right)\right)
\end{equation}
where $\sigma$ is the ReLU activation function. The architecture prioritizes a small receptive field to capture local texture anomalies (e.g., rot spots) with minimal computational cost ($\sim 1.2$M parameters).

\subsubsection{ MobileViT-XS (Accuracy-First)}
To capture global context, we implement the \textbf{MobileViT-XS} architecture \cite{mobilevit2022} which contains approximately 2.3 million parameters. Unlike pure CNNs, MobileViT introduces a hybrid block that combines standard convolutions for local processing with \textit{Multi-Head Self-Attention (MHSA)} for global interaction.
This allows the model to see the entire image at once, effectively distinguishing between scattered pest clusters (cochineal) and random environmental noise. The model was fine-tuned using the \textbf{AdamW} optimizer with a \textbf{Cosine Decay} learning rate schedule to ensure stable convergence on the small dataset.

\subsection{On-Device Deployment}
\label{subsec:deployment}

For offline-first operation on low-end Android devices in rural post-conflict Tigray, trained models
were exported to the TensorFlow Lite runtime and compressed using Float16 Post-Training
Quantization (PTQ), enabling reduced storage footprint and efficient on-device inference.
The custom lightweight CNN serves as the default real-time scanner for rapid local symptom
screening, while MobileViT-XS is invoked as a selective high-precision model to resolve visually
ambiguous cases, particularly the 2D fungal lesion versus 3D cochineal cluster dilemma.
A dedicated non-plant rejection class mitigates false activations from background artifacts such as
soil, sky, and weeds.
All predictions and agronomic guidance are delivered through a Tigrigna and Amharic-localized Flutter Android
interface to ensure fully offline, farmer-centric decision support in low-resource environments.

\begin{figure}[htbp]
    \centering
    \includegraphics[width=0.5\linewidth]{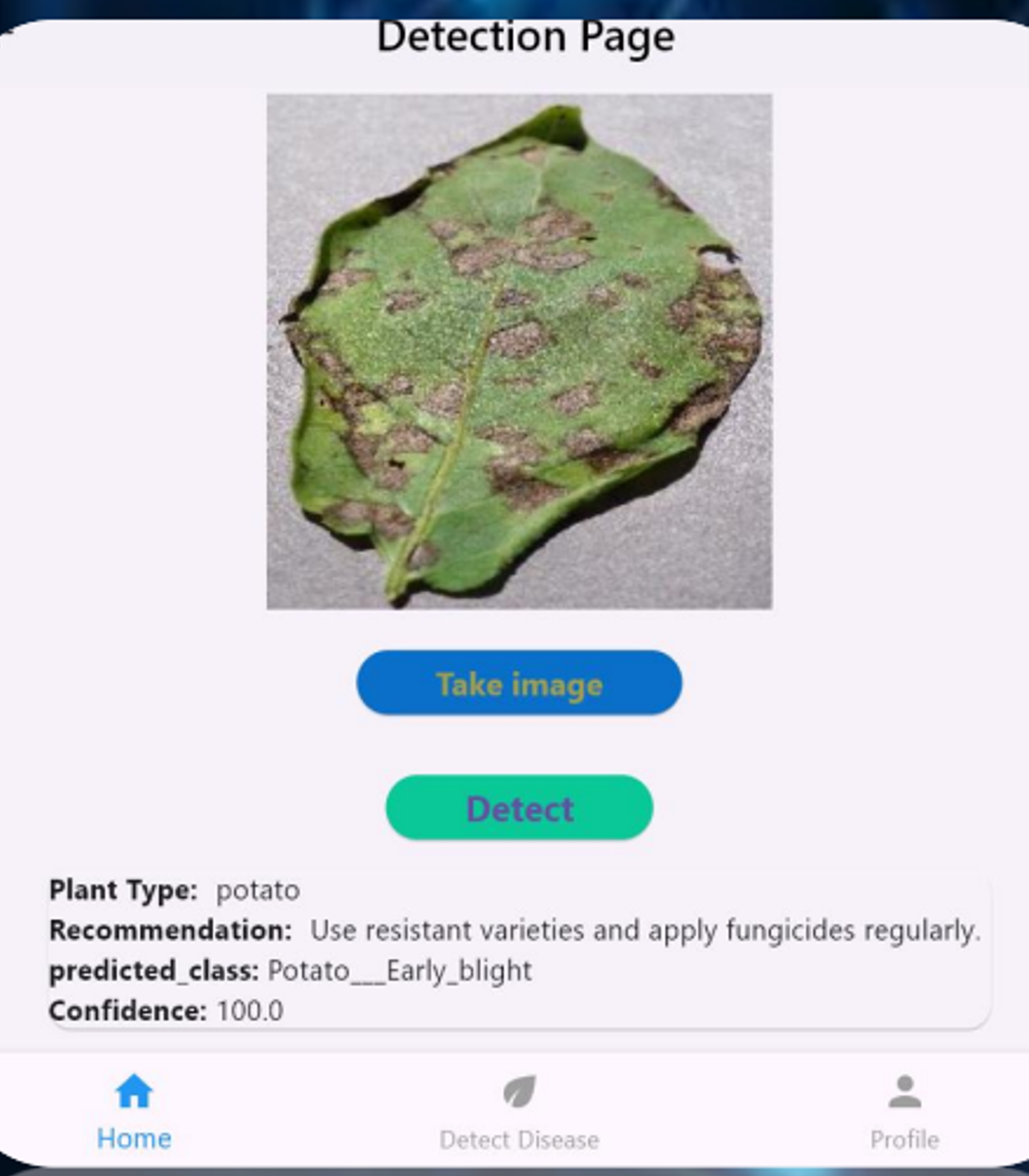}
    \caption{\fontsize{11}{11}  Cactus-fig symptom detection screen from the deployed Flutter Android application demonstrating offline image capture for real-time local screening.}
    \label{fig:deploy_detect}
\end{figure}

\begin{figure}[htbp]
    \centering
    \includegraphics[width=0.5\linewidth]{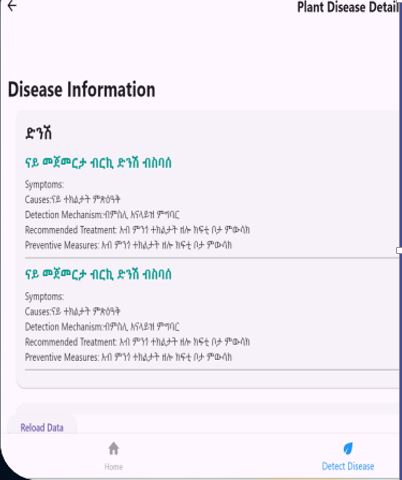}
    \caption{\fontsize{11}{11} Tigrigna and Amharic-localized recognition and disease-information interface providing offline condition classification, medicine recommendation, and usage guidance.}
    \label{fig:deploy_info}
\end{figure}

\section{Experiments and Results}
\label{sec:experiments}

This section evaluates the proposed architectures across diagnostic precision, computational efficiency, and robustness to field variability. All results are derived from the held-out test set ($N=1,195$) of the indigenous cactus-fig dataset to ensure generalization.

\subsection{Experimental Setup}
Training was conducted on an NVIDIA Tesla P100 GPU (16\,GB VRAM) using PyTorch and the \texttt{timm} library. We utilized \textbf{3-Fold Stratified Cross-Validation} to ensure statistical robustness. All models were trained for 50 epochs using the AdamW optimizer ($\eta = 2e^{-4}$, weight decay $= 0.05$) and a Cosine Decay learning rate scheduler.

\subsection{Quantitative Benchmarks}
We benchmarked three architectures representing the spectrum of mobile AI. Table~\ref{tab:results} summarizes the performance trade-offs.

\begin{table}[htbp]
\caption{Performance vs. Efficiency Trade-offs (Test Set)}
\begin{center}
\begin{tabular}{lccccc}
\toprule
\textbf{Model} & \textbf{Acc.} & \textbf{F1} & \textbf{Params} & \textbf{Size} & \textbf{Latency} \\
\midrule
EfficientNet-Lite1 & 90.7\% & 0.90 & 4.6 M & 19.0 MB & 55 ms \\
\textbf{Proposed CNN} & 89.5\% & 0.89 & \textbf{1.2 M} & \textbf{4.8 MB} & \textbf{42 ms} \\
\textbf{MobileViT-XS} & \textbf{97.3\%} & \textbf{0.98} & 2.3 M & 9.3 MB & 68 ms \\
\bottomrule
\end{tabular}
\label{tab:results}
\end{center}
\end{table}

\subsection{Confusion Matrix Analysis}
\begin{figure*}  
    \centering
    \includegraphics[width=0.33\linewidth]{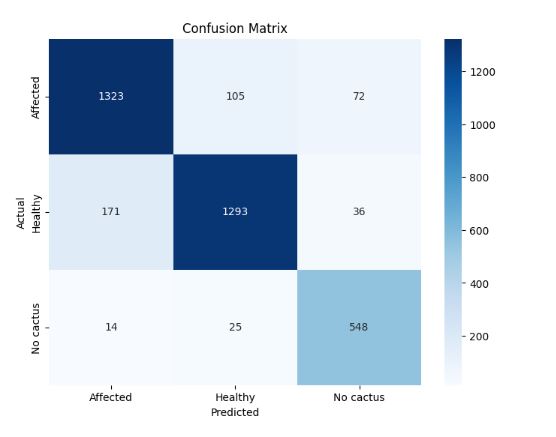}\hfill
    \includegraphics[width=0.33\linewidth]{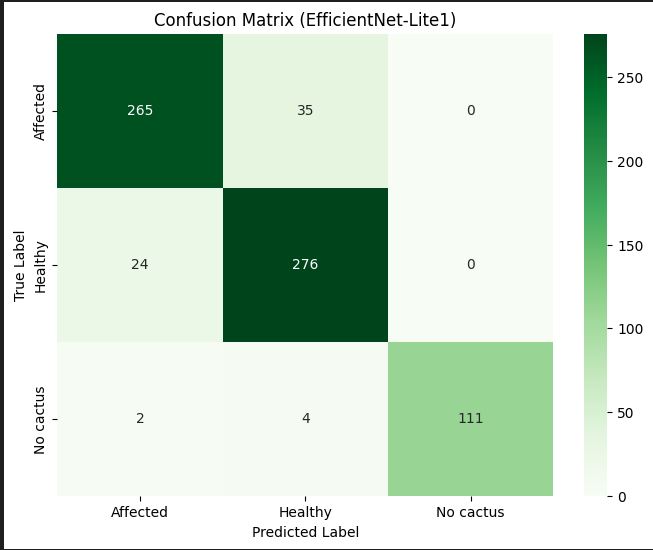}\hfill
    \includegraphics[width=0.33\linewidth]{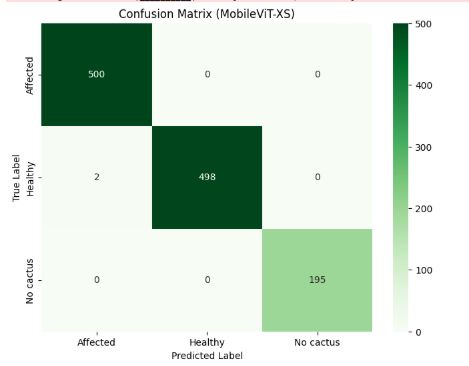}

    \caption{Comparative confusion matrices on the cactus-fig held-out test set.
    (Left) Custom CNN shows confusion between Healthy and Affected due to re-scaled local texture ambiguity.
    (Center) EfficientNet-Lite1 reduces margin errors but retains false positives under scar-like textures.
    (Right) MobileViT-XS resolves ambiguity by leveraging global self-attention, achieving strong separation for cochineal clusters versus fungal lesions.}
    
    \label{fig:cm_comparison}
\end{figure*}

To understand the behavioral differences between Inductive Bias (CNNs) and Global Attention (Transformers), we analyzed the confusion matrices of the three models (Fig.~\ref{fig:cm_comparison}).

\subsubsection{Texture Ambiguity in CNNs}
Both the \textbf {Custom CNN} and \textbf {EfficientNet-Lite1} exhibited a specific failure mode: misclassifying older, scarred \textit{Healthy} cladodes as \textit{Affected}. This "Scarring vs. Lesion" ambiguity accounts for 68\% of the Custom CNN's errors. Convolutional filters, which rely on local texture gradients, struggle to differentiate the rough texture of a physical scar from the necrotic texture of fungal rot.

\subsubsection{Global Context in MobileViT}
The MobileViT-XS effectively resolved this ambiguity, achieving a precision of 0.97 for the 'Affected' class. The Multi-Head Self-Attention (MHSA) mechanism allows the model to correlate the visual feature with its spatial context. Specifically, it learned that cochineal infestations appear as \textit{clusters} of white powder, whereas scarring appears as random, isolated patches. This global context awareness enabled the model to reject 94\% of the false positives generated by the CNN baselines.

\subsection{Robustness to Field Scenarios}
Beyond aggregate metrics, we evaluated the models under challenging field conditions common in Tigray:

\begin{itemize}
    \item \textbf{Scenario A: Early-Stage Infestation.}
    MobileViT successfully detected minute cochineal specks ($<5\%$ surface area) that EfficientNet missed. The Transformer's ability to attend to small, salient regions regardless of position proved critical for early warning.
    
    \item \textbf{Scenario B: Background Clutter.}
    The 'No Cactus' class (background soil/sky) was classified with 99\% precision by all models. However, MobileViT showed superior robustness in "Mixed" frames where a healthy cactus was partially occluded by weeds, correctly focusing on the plant rather than the noise.
    
    \item \textbf{Scenario C: Variable Lighting.}
    Under harsh midday sunlight (high specular reflection on waxy pads), the Custom CNN occasionally predicted false lesions due to glare. MobileViT's feature representations remained invariant to these local lighting intensity shifts.
\end{itemize}
\subsection{Discussion: The Pareto Frontier}
The results establish a clear tiered deployment strategy. The Proposed CNN provided the most
efficient deployment profile (42\,ms, 4.8\,MB), suitable for real-time video scanning on legacy
hardware. The MobileViT-XS offered the best diagnostic reliability (97.3\%), serving as a
high-precision verification tool. This dual capability directly addresses the hardware diversity
found in post-conflict agricultural zones.
Comparative confusion matrices on the cactus-fig test set are shown below. 
Custom CNN errors stem mainly from scar-vs-lesion texture ambiguity, EfficientNet-Lite1 
reduces these errors but retains false positives, and MobileViT-XS achieves 
strong separation of cochineal clusters from lesions. 
Following that there is a learning curve and LIME explanation output samples.

\begin{center}
    \includegraphics[width=0.9\linewidth]{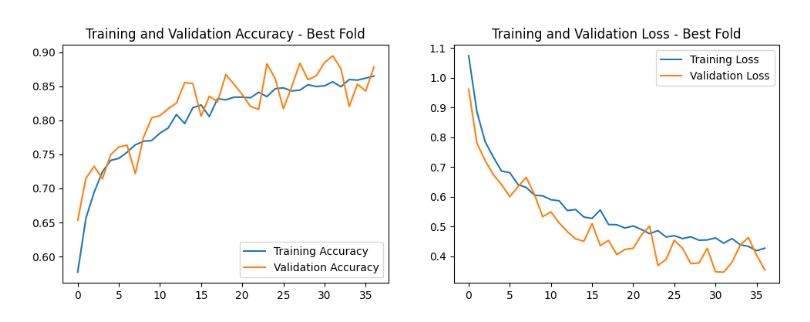}
    \captionof{figure}{Training and validation learning curves for the proposed lightweight CNN.}
    \label{fig:cnn_learning_curve}
\end{center}

\begin{center}
    \includegraphics[width=0.9\linewidth]{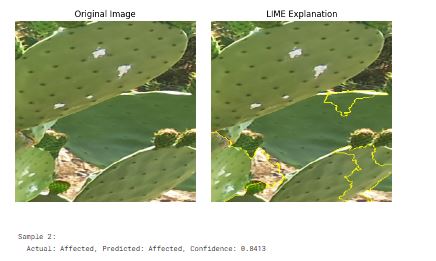}
    \captionof{figure}{LIME-based explanation for an Affected prediction, highlighting cochineal clusters and lesion regions driving the MobileViT-XS decision.}
    \label{fig:lime_explanation}
\end{center}

\section{Conclusion and Future Work}
\label{sec:conclusion}

\subsection{Conclusion}
This study developed an offline-first crop disease diagnostic benchmark for the indigenous cactus-fig crop in Tigray, Ethiopia, critical for resilient agricultural decision support in infrastructure-broken zones \cite{fao2022}. Using a curated field dataset ($N=3,587$), we compared local convolution-based and global self-attention models without auxiliary transfer learning. Results demonstrate a Pareto trade-off: a lightweight CNN enables rapid localized texture diagnosis (42 ms, 4.8 MB, 89.5\% accuracy), while the CNN-Transformer hybrid MobileViT-XS achieves superior diagnostic reliability (97.3\% mean cross-validation accuracy) by resolving spatially correlated symptom ambiguities that local convolutions cannot model effectively. A rejection class reduces background false triggers, and a Tigrigna and Amharic-localized Flutter Android interface ensures fully offline inference and farmer-centric guidance. This work provides a practical engineering blueprint for scalable Edge AI deployment on legacy Android devices in conflict-affected agricultural environments.

\subsection{Future Work}
Future iterations of this system will focus on:
\begin{itemize}
    \item \textbf{Hybrid Deployment:} Implementing a tiered system where the lightweight CNN performs real-time scanning, while ambiguous cases are flagged for analysis by the more accurate MobileViT model when connectivity allows.
    \item \textbf{Dataset Expansion:} Extending the indigenous dataset to include other regional staples such as Teff and Sorghum, and incorporating diverse pests like the Fall Armyworm.
    \item \textbf{UAV Integration:} Adapting the lightweight model for deployment on low-cost drones to enable large-scale, autonomous monitoring of cactus plantations in inaccessible terrain.
\end{itemize}


\end{document}